\documentclass[sigconf,authorversion,nonacm]{acmart}
\AtBeginDocument{%
  \providecommand\BibTeX{{%
    \normalfont B\kern-0.5em{\scshape i\kern-0.25em b}\kern-0.8em\TeX}}}

\setcopyright{acmcopyright}
\copyrightyear{2022}
\acmYear{2022}
\acmDOI{XXXXXXX.XXXXXXX}

\acmConference[MM '22]{In Proceedings of the 30th
ACM International Conference on Multimedia}{October 10--14,
  2022}{Lisbon, Portugal}
\acmBooktitle{Woodstock '18: ACM Symposium on Neural Gaze Detection,
 June 03--05, 2018, Woodstock, NY} 
\acmPrice{15.00}
\acmISBN{978-1-4503-XXXX-X/18/06}

\usepackage{multirow}

\begin{document}

\title{OS-MSL: One Stage Multimodal Sequential Link Framework for Scene Segmentation and Classification}

\author{Ye Liu}
\authornote{Both authors contributed equally to this research.}
\affiliation{%
  \institution{Tencent YouTu Lab}
  \city{Shanghai}
  \country{China}
}
\email{rafelliu@tencent.com}

\author{Lingfeng Qiao}
\authornotemark[1]
\affiliation{%
  \institution{Tencent YouTu Lab}
  \city{Shanghai}
  \country{China}
}
\email{leafqiao@tencent.com}

\author{Di Yin}
\affiliation{%
  \institution{Tencent YouTu Lab}
  \city{Shanghai}
  \country{China}
}
\email{endymecyyin@tencent.com}

\author{Zhuoxuan Jiang}
\affiliation{%
  \institution{Tencent YouTu Lab}
  \city{Shanghai}
  \country{China}
}
\email{jzhx1211@gmail.com}

\author{Xinghua Jiang}
\affiliation{%
  \institution{Tencent YouTu Lab}
  \city{Hefei}
  \country{China}
}
\email{clarkjiang@tencent.com}

\author{Deqiang Jiang}
\affiliation{%
  \institution{Tencent YouTu Lab}
  \city{Hefei}
  \country{China}
}
\email{dqiangjiang@tencent.com}

\author{Bo Ren}
\affiliation{%
  \institution{Tencent YouTu Lab}
  \city{Hefei}
  \country{China}
}
\email{timren@tencent.com}

\renewcommand{\shortauthors}{Ye Liu and Lingfeng Qiao, et al.}

\begin{abstract}

Scene segmentation and classification (SSC)  serve as a critical step towards the field of video structuring analysis. Intuitively, jointly learning of these two tasks can promote each other by sharing common information. However, scene segmentation concerns more on the local difference between adjacent shots while classification needs the global representation of scene segments, which probably leads to the model dominated by one of the two tasks in the training phase. In this paper, from an alternate perspective to overcome the above challenges, we unite these two tasks into one task by a new form of predicting shots link: a link connects two adjacent shots, indicating that they belong to the same scene or category. To the end, we propose a general One Stage Multimodal Sequential Link Framework (OS-MSL) to both distinguish and leverage the two-fold semantics by reforming the two learning tasks into a unified one. Furthermore, we tailor a specific module called DiffCorrNet to explicitly extract the information of differences and correlations among shots. Extensive experiments on a brand-new large scale dataset collected from real-world applications, and MovieScenes are conducted. Both the results demonstrate the effectiveness of our proposed method against strong baselines.


\end{abstract}




\maketitle

\section{Introduction}

With massive video data generated everyday, video structuring and understanding are significant work for media resource management. For example in a video clips search engine, long videos need to be pre-processed and segmented into different topics for later fast retrieval. 
Therefore, long video segmentation and topic classification are fundamental techniques for various senior media applications, which, however, is still less-studied in the community.


\begin{figure}[t]
  \centering
  \includegraphics[width=0.9\linewidth]{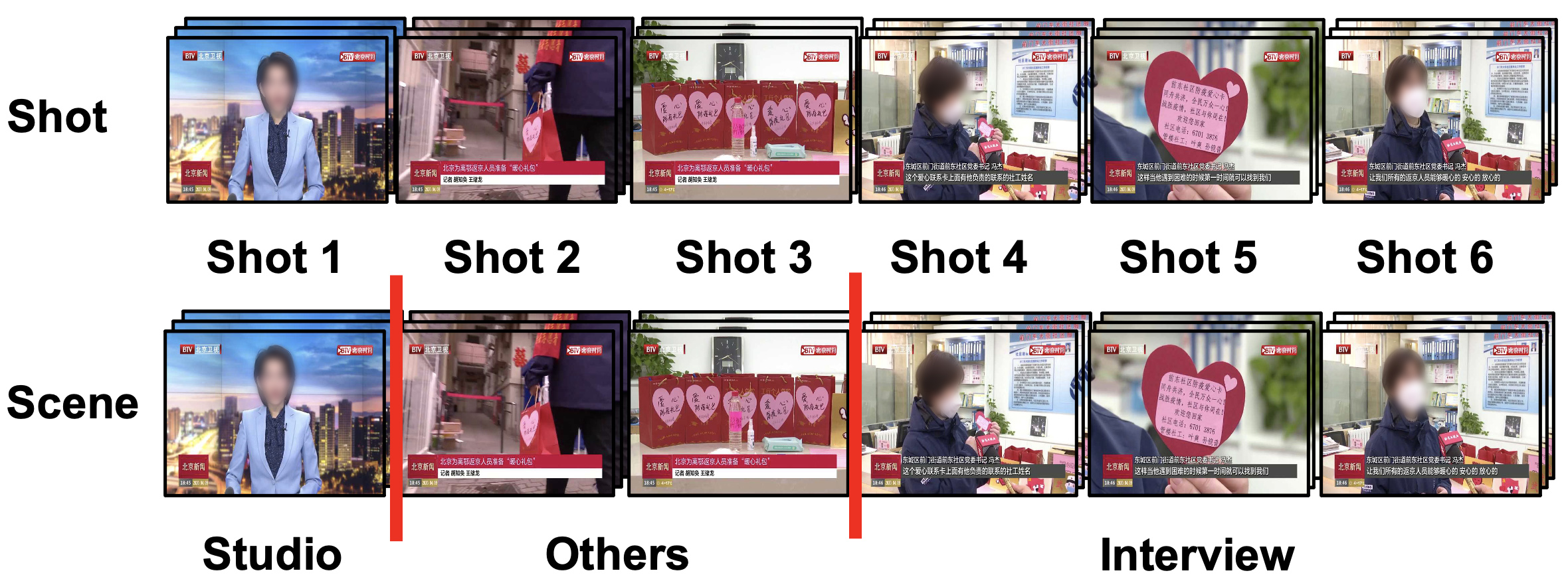}
  \caption{Example of shot and scene in a long video. Similar frames constitute shots and related shots constitute scenes.}
  \label{FIG: shot scene}
\end{figure}


As known, usually we have two types of video segments: shot and scene for different scales. \textit{\textbf{Shot}} is a set of successive frames taken without interruption by a single camera \cite{boccignone2005foveated}, and \textit{\textbf{Scene}} is the combination of adjacent shots which are related with similar background or environment. As illustrated in Figure \ref{FIG: shot scene}, the first row shows that we can segment six shots from a long video where each shot is composed of frames shot by the same camera. The second row shows the partition of similar shots which constitute scenes, e.g. \textit{studio} and \textit{interview}. The red lines represent the segment boundaries. 
To this end, this paper conducts a pioneer study to investigate the problem that how to segment long videos and classify each segment in a joint manner, and we name it as Scene Segmentation and Classification (SSC).


As mentioned above, video segmentation at the shot level is a relatively simple task, and adjacent frames can be easily identified to different shots or the same due to the distinct low-level visual features~\cite{cotsaces2006video,smeaton2010video}. However, the scene segmentation (SS) is pretty tough because those shots that belong to one scene could be totally visually dissimilar but they are semantically consistent, which requires high-level understanding to the content. For example, in a scene of multi-party dialogue, shot cut often happens to give close-up for the one who is speaking. Therefore, low-level visual algorithms may not work well and the ability to sufficiently understand the correlation between shots is essential~\cite{chasanis2008scene}.

Besides video segmentation for structuring, classification for each segment is also a crucial step for video understanding. Intuitively,  SS focuses more on the local difference between two consecutive shots, as there should be a boundary if these two shots are not similar. In contrast, scene classification needs the global information between shots, as multiple shots in one scene has the common category label. However, most existing work considers to extract the hierarchical global representations  ~\cite{carreira2017quo,feichtenhofer2019slowfast}, where the local sequence knowledge learned by scene segmentation has not been well leveraged, leading to a worse performance.  

\begin{figure}[t]
  \centering
  \includegraphics[width=0.7\linewidth]{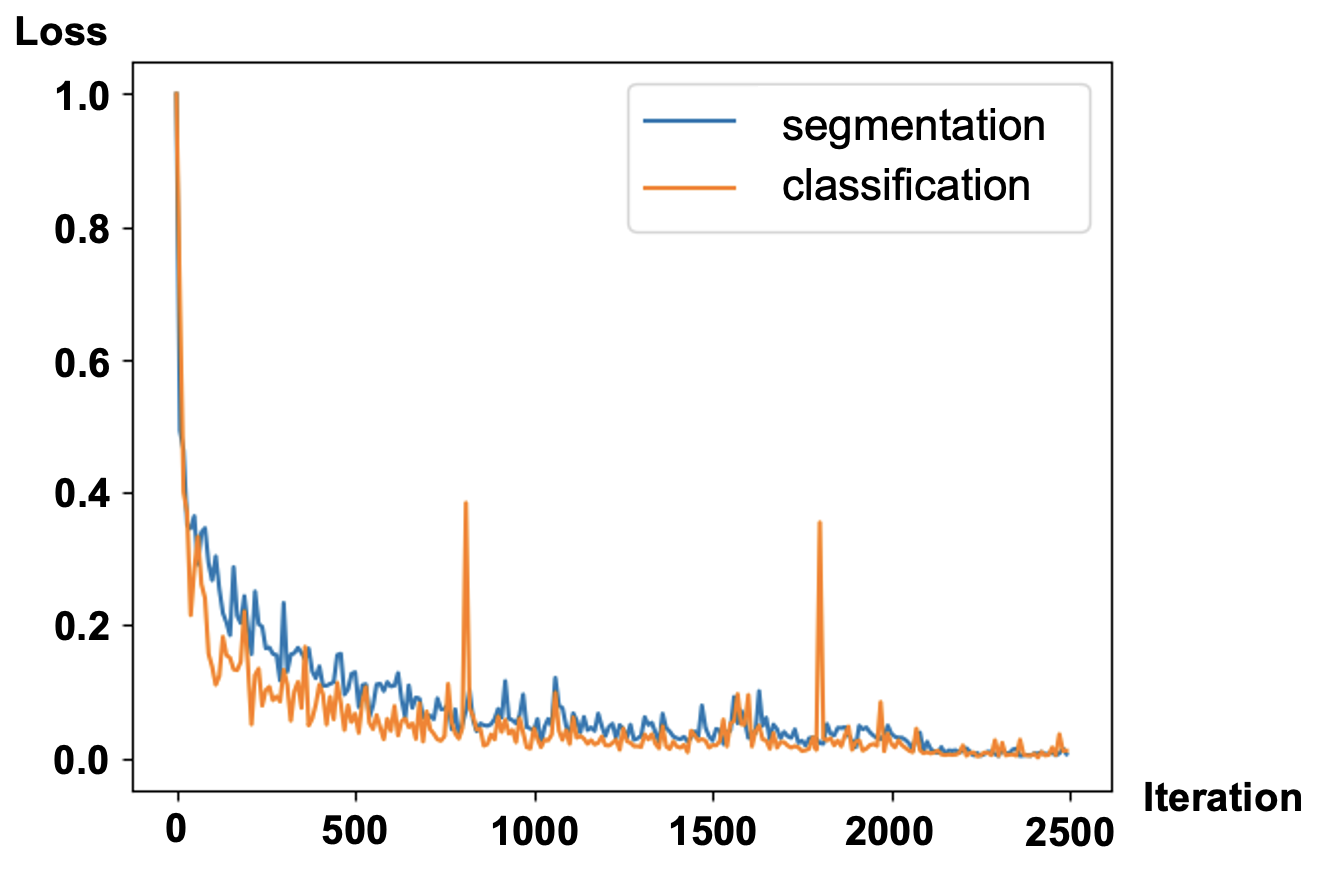}
  \caption{The learning curve of multi-task model for SSC.}
  \label{FIG:learning curve introduction}
  \vspace{-4ex}
\end{figure}

To handle the segmentation and classification problems together, a natural way is to utilize the two-stage method. In detail, segmentation task is first performed to obtain scenes and then the scenes are labelled with video classification algorithm. However, the method may encounter error accumulation and miss the relation information between the two tasks. Another solution is to apply multi-task learning strategy. Nevertheless by our experiments and analysis, we observe that simply building a multi-task learning model could not achieve ideal performance. We use two separate loss functions and train the two tasks in a vanilla manner \cite{caruana1997multitask}. Figure \ref{FIG:learning curve introduction} shows the learning curve of two losses during training phase by 2,500 iterations. It can be found that the classification curve is oscillating while the segmentation curve is relatively stable, which indicates that the two tasks have different gradient descending directions and multi-task learning is probably not a superior solution. The details of analysis could refer to the Experiments section.

\begin{figure}[t]
  \centering
  \includegraphics[width=0.9\linewidth]{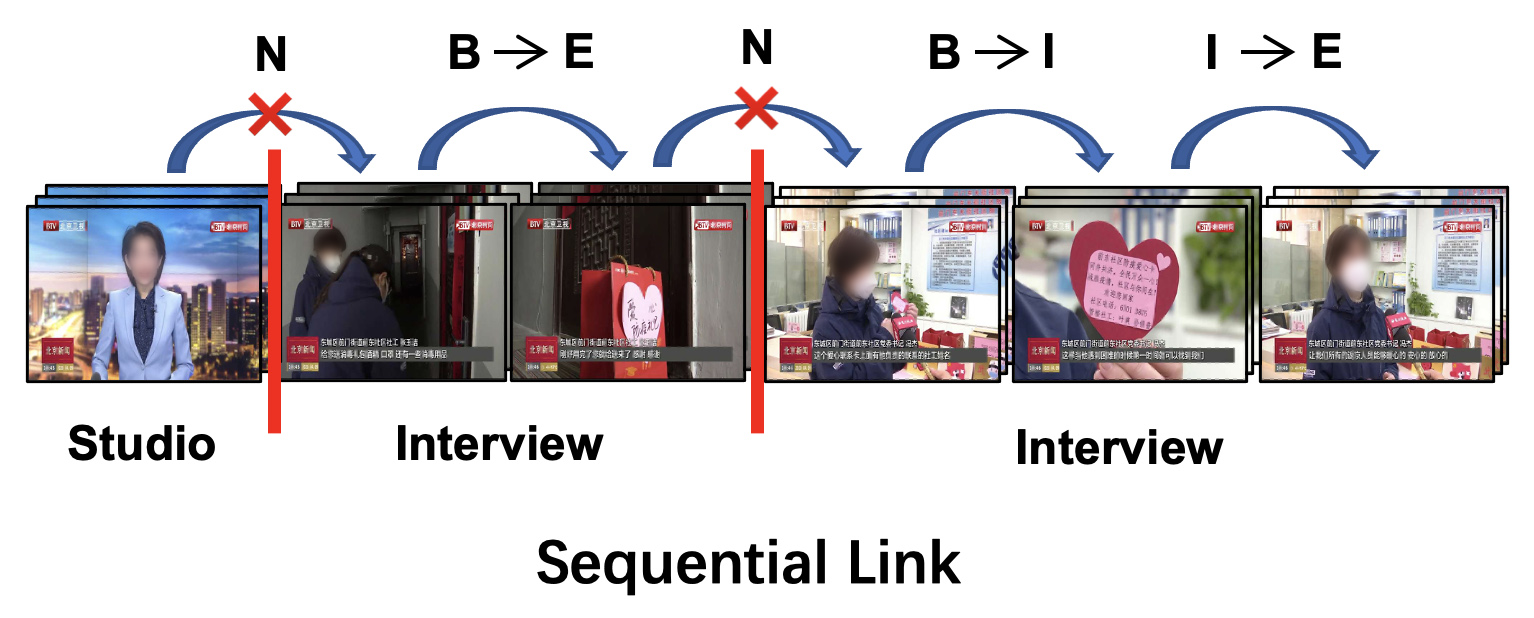}
  \caption{Schematic diagrams of Sequential Link framework. In our framework, segmentation and classification tasks are unified into one-stage task by predicting shots link.}
  \label{FIG: method compare}
\end{figure}

To overcome the above challenges, we tackle the SS and SSC problems in a new perspective. We propose to decompose long videos into two elements, namely \textit{shot} and \textit{link}. 
A link connects a pair of adjacent shots, indicating that they belong to the same scene. Under the above definitions, as illustrated in Figure \ref{FIG: method compare}, the task of SS is reformed to predict whether there is a link between current shot and next shot. In detail, by enumerate all the possible links, we have
a 5-classes link label set: $\{$B->I, I->I, I->E, B->E, N$\}$,  where B is Beginning, I is Intermediate, E is End, and N means there is no link between these two shots. For example, suppose  a scene consists of multiple shots, thus the beginning shot should be linked with some/zero intermediate shots, ended with an end shot.

SSC is an extension of SS, whose label set is denoted as $C \times \{$B->I, I->I, I->E, B->E, N$\}$ where $C$ is the number of classes. Next, multimodal information, visual and audio, are leveraged to extract powerful representation of shots~\cite{sundaram2000video}. It should be pointed out that text modality is not used because the linguistic contents in video are little related to SSC tasks. More discussions about these modalities could refer to the Experiments section. Then, to better model the difference and correlation between shots which is much important for the SSC, a tailored network (called DiffCorrNet) is proposed for particularly extracting shot features. At last, a transformer structure and a conditional random fields (CRF) module are utilized to make predictions on link tags. We name the proposed model with \textbf{O}ne \textbf{S}tage \textbf{M}ultimodal \textbf{S}equential \textbf{L}ink Framework (OS-MSL).

Our method  provides a general framework for other similar video sequential modeling problems, e.g., action, activity or event recognition, which can be categorized to  a unified definition: find a boundary between two consecutive segments/clips with auxiliary information such as label, key frame, and caption.

In addition, we contribute a new large-scale dataset called \textbf{TI-News} for scene understanding community. Existing related datasets, such as MovieScenes \cite{rao2020local}, only focus on movie domain and they solely offer the SS labels. To expand the diversity of video domains and broaden the application scenarios, we collected real world news videos with sufficient SSC annotations. TI-News can enrich the dataset resources in scene understanding community and can promote the researches and applications in this area.




We summarize the main contributions as follows:

\begin{itemize}





\item We provide an alternate perspective to define the SS and SSC problems to a new  linking form, and propose a one stage multimodal sequential link framework, which can unify multiple tasks in SSC  into a one-stage task by predicting shots link.


\item We propose a feature extraction network DiffCorrNet to simultaneously extract the difference and correlation between adjacent shots.


\item We construct a new dataset TI-News which include hundreds news videos with both segment and category labels.

\item Extensive experiments on two datasets TI-News and MovieScenes demonstrate the effectiveness of our method and achieve state-of-the-art results. 


\end{itemize}

\begin{figure*}[h]
  \centering
  \includegraphics[width=0.9\linewidth]{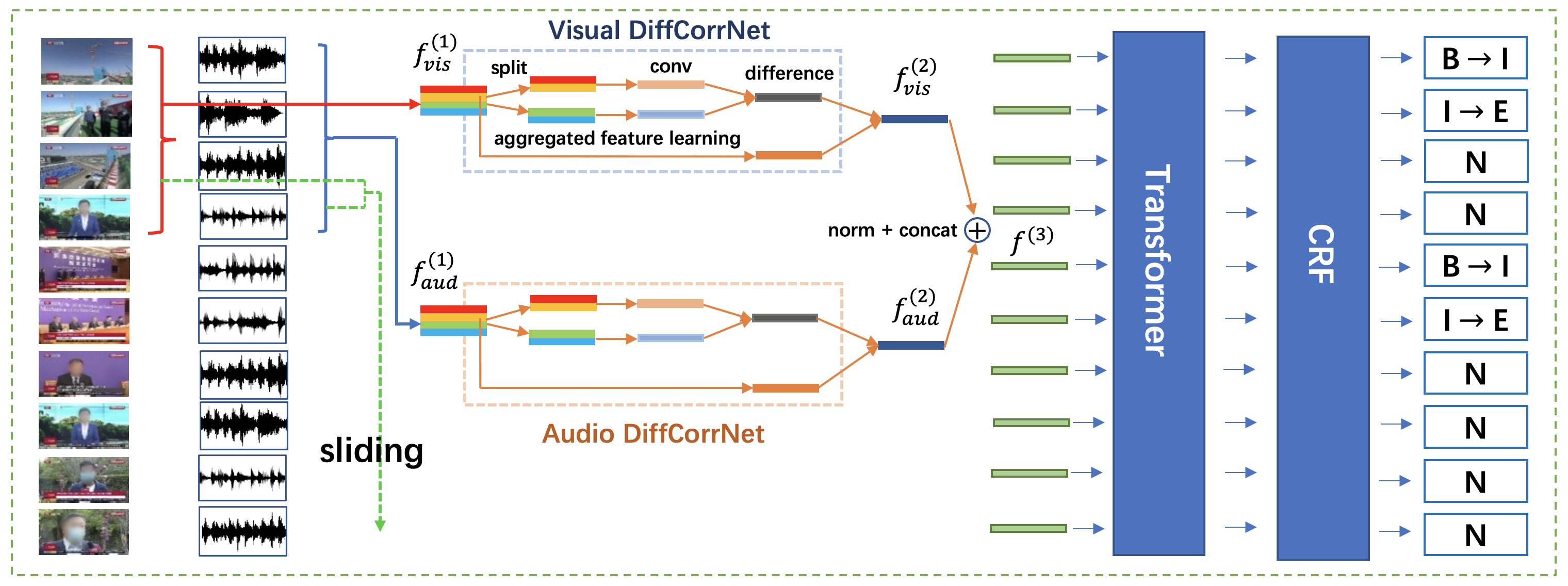}
  \caption{The schematic diagram of our OS-MSL algorithm.}
  \label{FIG:algorithm}
\end{figure*}

\section{Related Work}
\subsection{Video Segmentation}


Traditionally, graph models are widely proposed to group shots where each shot is regarded as node in a graph and segmentation is implemented by graph cut algorithms~\cite{rasheed2005detection,yeung1998segmentation,sidiropoulos2011temporal}. Later graph convolution network is applied to this line \cite{pei2021video}. On the other hand, similarity measurement, such as siamese network~\cite{baraldi2015deep}, and shot clustering methods are proposed for video segmentation~\cite{rotman2016robust,rotman2018optimally,rotman2020learnable}. Recently, with a large-scale dataset released \cite{huang2020movienet}, end-to-end models are designed to detect boundaries directly. LGSS~\cite{rao2020local} raises B-Net to identify boundaries and uses multimodal features, and ShotCoL \cite{chen2021shot} leverages contrastive learning. However, their work performed a 2-classes classification which refers to whether the shot should be segmented or not, ignoring richer relation information between shots within a scene. Different from their work, we reform SS task to a link form, and  explicitly reveal the sequential relation between shots under a 5-classes richer link labels, which can promote the SS task.


\subsection{Video Classification}
Existing video classification works are mostly for single task and usually applied to short videos or video snippets segmented from long videos, such as hand-crafted feature engineering method~\cite{wang2013dense}, deep learning based Two-stream CNNs \cite{simonyan2014two, wang2018temporal}, 3D-CNN \cite{tran2015learning}, and vision transformer based method~\cite{arnab2021vivit}. As to in a joint task for long videos, existing classification works may suffer from a decreased performance of the preceding segmentation in a pipeline way. This paper tries to alleviate this issue in an ingenious way. 

\subsection{Video Segmentation and Classification}
It is easily confusing that our SSC task is similar to the Temporal Action Detection (TAD) task, because both conduct segmentation and classification on long videos. Proposal generation and action classification are the two main tasks in TAD, and two-stage or multi-task methods are proposed with particular techniques such as sliding window \cite{shou2016temporal}, anchor-based method~\cite{gao2017turn}, Boundary Sensitive Network~\cite{lin2018bsn} and transformer structure with an end-to-end set-prediction-based detector~\cite{liu2021end}. However, the segmentation in our SSC task should process the whole video rather than only detecting actions, which is different from TAD. Specially, the two-stage and multi-task based algorithms in TAD field are able to handle SSC problem, while the two schemes have limitations which are mentioned in introduction section. Different with them, our Sequential Link framework can overcome their shortcomings and is proved effective in SSC experiments.



\section{Problem Definition}

SSC includes two tasks, i.e. scene segmentation and scene classification. It needs to partition the scene segments on the basis of the predefined scene categories. SS aims to extract all the segments satisfying the scene definition from the entire video. Different scene segments need to cover whole video and they must be exclusive, without overlapping. Besides, the goal of scene classification is to assign correct scene label to each segment. 

Because scenes are composed of consecutive shots, to simplify the scene problem, algorithms for SSC can be designed on the basis of the shot segmentation results. For an input video, firstly, it is segmented into several shots by shot algorithm \cite{souvcek2020transnet}. The shots are denoted as $S_{shot}=\left \{a_{1},a_{2},\cdots ,a_{n_{1}} \right \}$, where $n_{1}$ is the number of shots and the subscript numbers are indexed by time order. Scene algorithm takes efforts to group these shots into $n_{2}$ scenes $S_{scene}=\left \{b_{1},b_{2},\cdots ,b_{n_{2}} \right \}$. Each scene $b_{i}$ consist several consecutive shots. The index of the start shot among them is denoted as $a_{s_{i}}$ and similarly, $a_{e_{i}}$ indicates the end shot. As a result, $b_{i} = \left \{a_{s_{i}},a_{s_{i}+1},\cdots,a_{e_{i}} \right \}$. Because each shot needs to be assigned to exactly one scene, it should be satisfied that $s_{i+1} = e_{i}+1, i=1,2,\cdots,n_{2}-1$. Furthermore, each scene $b_{i}$ needs to be classified. The label prediction of $b_{i}$ is denoted as $l_{i},i=1,2,\cdots,n_{2}-1$. To sum up, $S_{scene}=\left \{b_{1},b_{2},\cdots ,b_{n_{2}} \right \}$ is the result of scene segmentation. $L_{scene}=\left \{l_{1},l_{2},\cdots ,l_{n_{2}} \right \}$ is the result of scene classification, where $l_{i}$ is the class label of scene $b_{i}$.

\section{Methodology}


The schematic diagram of our OS-MSL is shown in Figure \ref{FIG:algorithm}. The main idea is extracting the multimodal feature of shots and then using sequential link framework with link tagging to solve segmentation and classification simultaneously. The architecture of OS-MSL consists of 3 parts. Firstly, unimodal features of shots are extracted. Then, features of different modalities are fused and the multimodal representations of shots are obtained. Finally, a transformer followed by CRF is used to tag the shots so as to determine the results of segmentation and classification.

\subsection{Sequential Link Framework with Link Tagging}

We use a link tagging to describe the linking relations between adjacent shots. 
In SS task, the link label set is: $\{$B->I, I->I, I->E, B->E, N$\}$,  where "B": Beginning, "I": Intermediate,  "E": End, "N": no link. Therefore, e.g., 
a scene with 5 adjacent shots is tagged as $\{$B->I, I->I, I->I, I->E, N$\}$, which means the beginning shot is linked with 3 intermediate shots and ended with an ending shot that is not linked with the next scene.

SSC can be easily extended by SS using an expanded link label set with $5*C$ classes denoted as $C \times \{$B->I, I->I, I->E, B->E, N$\}$ where $C$ is he number of classes. For example, one \textit{Meeting} scene with 5 adjacent shots is tagged as: ${Meeting}_{B->I}$, ${Meeting}_{I->I}$, ${Meeting}_{I->I}$, ${Meeting}_{I->E}$ and ${Meeting}_{N}$. ${Meeting}_{B->I}$ stands for the beginning shot of a meeting scene, followed by an intermediate shot. ${Meeting}_{I->I}$ means the shot is an intermediate shot within a meeting scene, followed by another intermediate shot. ${Meeting}_{I->E}$ means the shot is an intermediate shot within a meeting scene, followed by an ending shot. ${Meeting}_{N}$ means the shot is an ending shot of a meeting scene, and there is no link between the  ending shot and the next scene.

\subsection{Unimodal Feature Extraction}

To assign shots with scene labels, the features of shots should be extracted. A shot, as a segment of the video, has its visual and audio stream signals. The multimodal signals can provide abundant information so that we need to extract the multimodal representation of the shots. We use the visual and audio modalities. Nevertheless, because there exists large gap between visual and audio signals, unimodal features should be obtained first.





For visual modality, ResNet-18 \cite{he2016deep} is used as the backbone of feature extractor. The visual feature is denoted as:
\begin{equation}
f_{vis}^{(1)}=F_{vis}\left ( s_{vis},\theta _{vis} \right )
\end{equation}
\noindent where $F_{vis}$ is the network architecture of ResNet-18, $\theta _{vis}$ is the parameters of $F_{vis}$ and $s_{vis}$ is the key frame of the shot. Currently, the frame at the middle time of the shot is chosen as the key frame.

Correspondingly, ResNet-VLAD \cite{xie2019utterance} is used as the backbone of audio stream. The audio feature is denoted as:
\begin{equation}
f_{aud}^{(1)}=F_{aud}\left ( s_{aud},\theta _{aud} \right )     
\end{equation}
\noindent where $F_{aud}$ is the network architecture of ResNet-VLAD, $\theta _{aud}$ is the parameters of $F_{aud}$ and $s_{aud}$ is the audio spectrogram of the shot. The spectrogram is obtained by preprocessing the audio wave, including the short-time Fourier transform (STFT) and Mel-scale filter banks.

\subsection{Shot Difference and Correlation Network}

The shot feature is used to fulfill two tasks. One is to decide whether a segmentation point is placed at the end of the shot. The other is to express the category information. In order to reinforce the feature representation, the local and global relations among adjacent shots are crucial. Therefore, DiffCorrNet is proposed to model the relations. For a certain shot $a_{j}$, consider its adjacent shots $a_{j-(k-1)}, a_{j-(k-2)},\dots,a_{j},$ $\dots,a_{j+k}$. The total $2k$ shots are used to generate the enhanced feature of $a_{j}$.

\begin{figure}[h]
  \centering
  \includegraphics[width=0.7\linewidth]{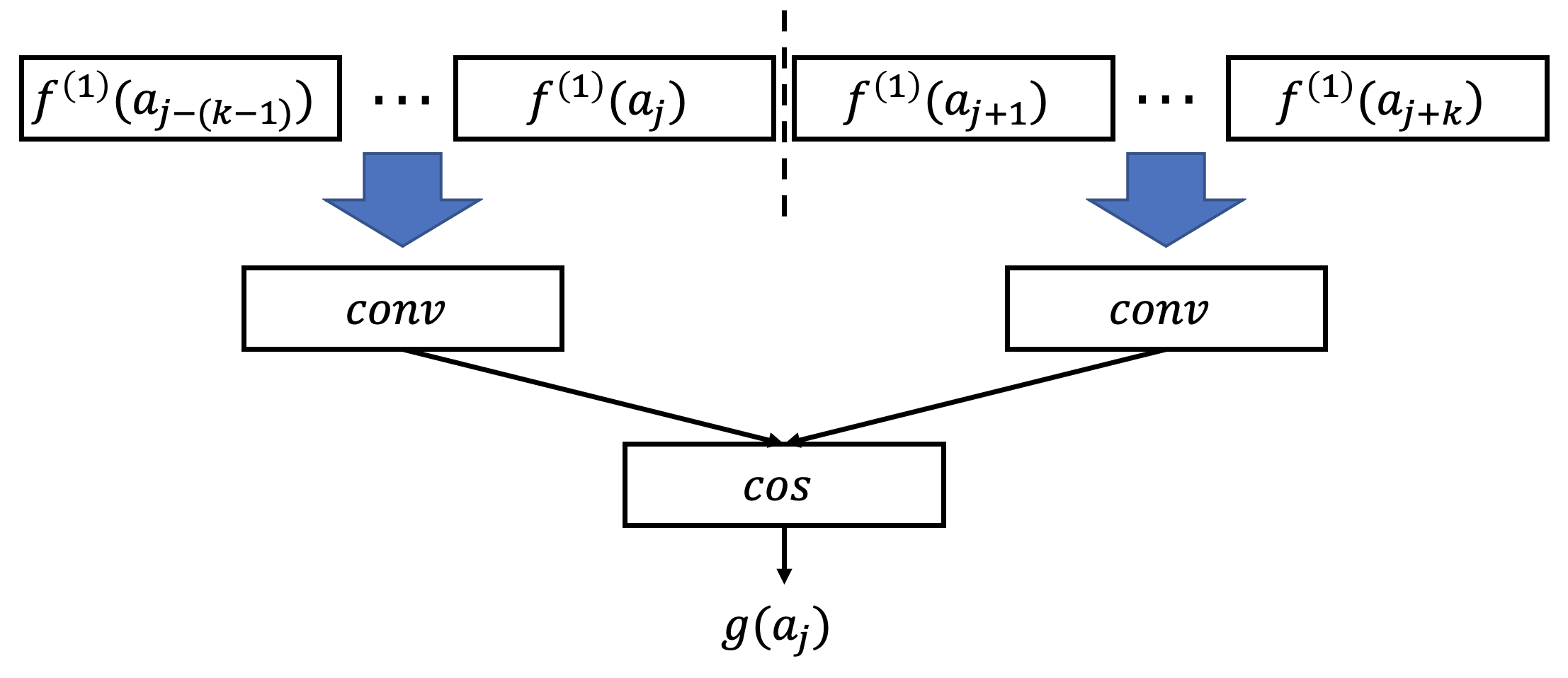}
  \caption{The network for learning boundary feature.}
  \label{FIG:boundary feature}
\end{figure}

At first, the boundary modeling is constructed by measuring the difference between the former $k$ shots and the latter $k$ shots. The difference can describe the confidence of whether the scene boundary is placed at the end of shot $a_{j}$. It is intuitional because if the former shots and the latter one are dissimilar, they likely belong to different scenes. Inspired by BNet module proposed in \cite{rao2020local}, we also use $cos$ metric to model the boundary information $g_{modal}$, where $modal\in\{vis,aud\}$. The model is shown in Figure \ref{FIG:boundary feature}.

\begin{figure}[h]
  \centering
  \includegraphics[width=0.7\linewidth]{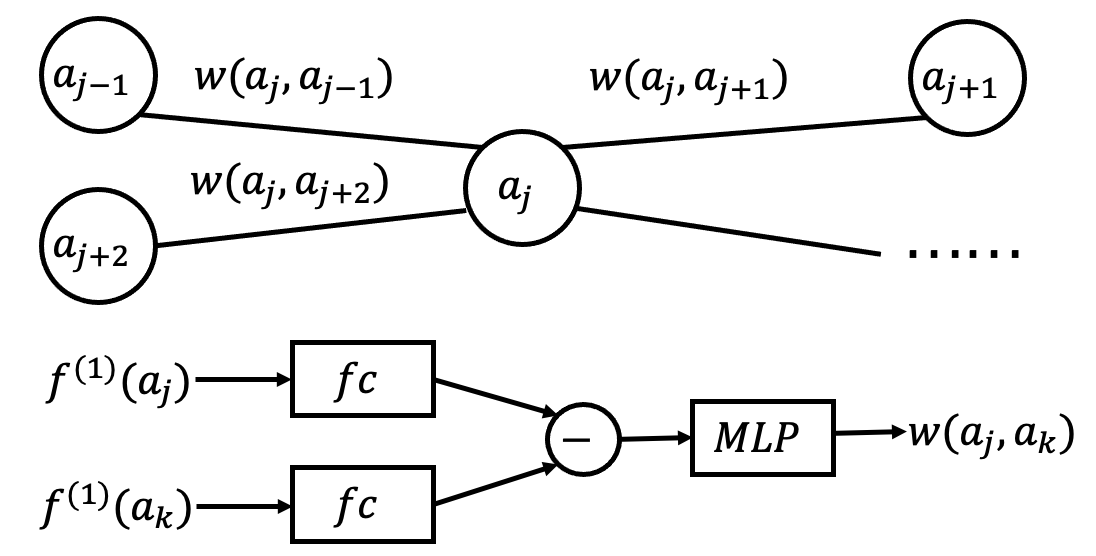}
  \caption{The network for learning aggregated feature.}
  \label{FIG:aggregated feature}
\end{figure}

To describe the global correlation between shots, the $f_{modal}^{(1)}$ of $2k$ shots are aggregated into a feature $h_{modal}$. Features of $2k-1$ neighboring shots are weighted by a shot attention mechanism and then summed to get the fused feature. The concept of this method is illustrated in Figure \ref{FIG:aggregated feature}.

$w(a_{j},a_{k})$ is denoted as the weight between shot $a_{j}$ and $a_{k}$. Concretely, the shot features are transformed by a fully connected network and then do subtraction to get a representation measuring the shot difference. Finally, a multilayer perceptron (MLP) is applied to get the weight.

After calculating all the weights between $a_{j}$ and its neighboring shots, the aggregated feature is obtained:

\begin{equation}
h(a_{j})=\sum_{i\in A(a_{j})}w(a_{j},a_{i})f^{(1)}(a_{i})
\end{equation}

\noindent where $A(a_{j})=\{j-(k-1), j-(k-2), \dots, j+k\} \setminus \{j\}$.

The purpose of this aggregated feature is to acquire the semantic information in adjacent shots with highly relation. The abundant information can be beneficial for better shot classification.

To sum up, the proposed DiffCorrNet including the sub networks mentioned above takes the $2k$ adjacent shots around shot $a_{j}$ into account and generates boundary feature  $g(a_{j})$ and aggregated feature $h(a_{j})$. The final unimodal features of $a_{j}$ are:

\begin{equation}
f_{modal}^{(2)}=concat(f_{modal}^{(1)},g_{modal},h_{modal})
\label{diffcornet}
\end{equation}

\noindent where $concat(\cdot)$ means the operation of feature concatenation. The parameters in DiffCorrNet are denoted as $\theta _{dc}$. Each modality has its own DiffCorrNet and the parameters in different DiffCorrNets are not shared.

By this means, DiffCorrNet can guide the extraction of shot feature for better scene segmentation and classification.

\subsection{Multimodal Feature Fusion}

Before feature fusing, batch normalization (BN) \cite{ioffe2015batch} is used to uniform the scales of different modalities, as the distributions and amplitudes of visual and audio signals are quite different. 

It is significant to fuse single-modality features to get the multimodal representation of the shots. In theory, there are several ways to fuse $f^{(2)}$, e.g. early fusion or late fusion. In OS-MSL, early fusion is adopted so that the transformer may take the multimodal feature into account.
As a result, the multimodal features of shots $f^{(3)}$ can be obtained by:

\begin{equation}
f^{(3)}=concat(BN(f_{vis}^{(2)}),BN(f_{aud}^{(2)}))
\label{after_concat}
\end{equation}
where the fused feature is obtained by concatenating all the provided unimodal features. This framework can be compatible with more modalities if necessary. 

\subsection{Task head}
Task head aims to complete the SS and SSC tasks. In this paper, a transformer \cite{vaswani2017attention}  followed by  conditional random fields (CRF) \cite{huang2015bidirectional} is chosen as the task head. 
Denote the head as $F_{head}$,  we give the shot label $y$ as:

\begin{equation}
y=F_{head}\left ( f^{(3)},\theta _{head} \right )
\end{equation}

\noindent where $\theta _{head}$ is the parameters of $F_{head}$.

The equations (1)-(6) represent the entire network of OS-MSL with trainable parameters $\theta _{vis},\theta _{aud},\theta _{dc},\theta _{head}$, which are trained by CRF loss \cite{huang2015bidirectional}. After training, each shot of a testing video can be classified by OS-MSL with $5\times C$ link labels. The final SSC results $\left \{b_{1},b_{2},\cdots ,b_{n_{2}} \right \}$ of segmentation boundaries and $\left \{l_{1},l_{2},\cdots ,l_{n_{2}} \right \}$ of category labels will be inferred by the predicted shots links $\left \{y_{1},y_{2},\cdots ,y_{n_{1}} \right \}$. In detail, consecutive shots or one shot with  link labels $\{$B->I, I->I, I->E, N$\}$ or $\{$B->E, N$\}$ or $\{$N$\}$ in category $c$ are treated as one scene, and its classification label is $c$.

\section{Experiments}

\subsection{Dataset}

\textbf{TI-News} To promote the video structuring research for news application, a new dataset, TI-News, is presented. It is one of the significant contributions in this paper. The dataset has over 500 news videos, including about 253,000 shots and 26,500 scenes. In training and validation phase, all the videos come from 8 news programs. In testing phase, 44 videos are selected for evaluation. 34 videos of them are chosen from aforementioned 8 programs. There is no overlap between the 34 testing videos and the training videos. Furthermore, 10 more videos collected from other 10 distinct programs are used for generalization testing. The detailed dataset splits is shown in supplementary material.

\begin{figure}[t]
  \centering
  \includegraphics[width=0.9\linewidth]{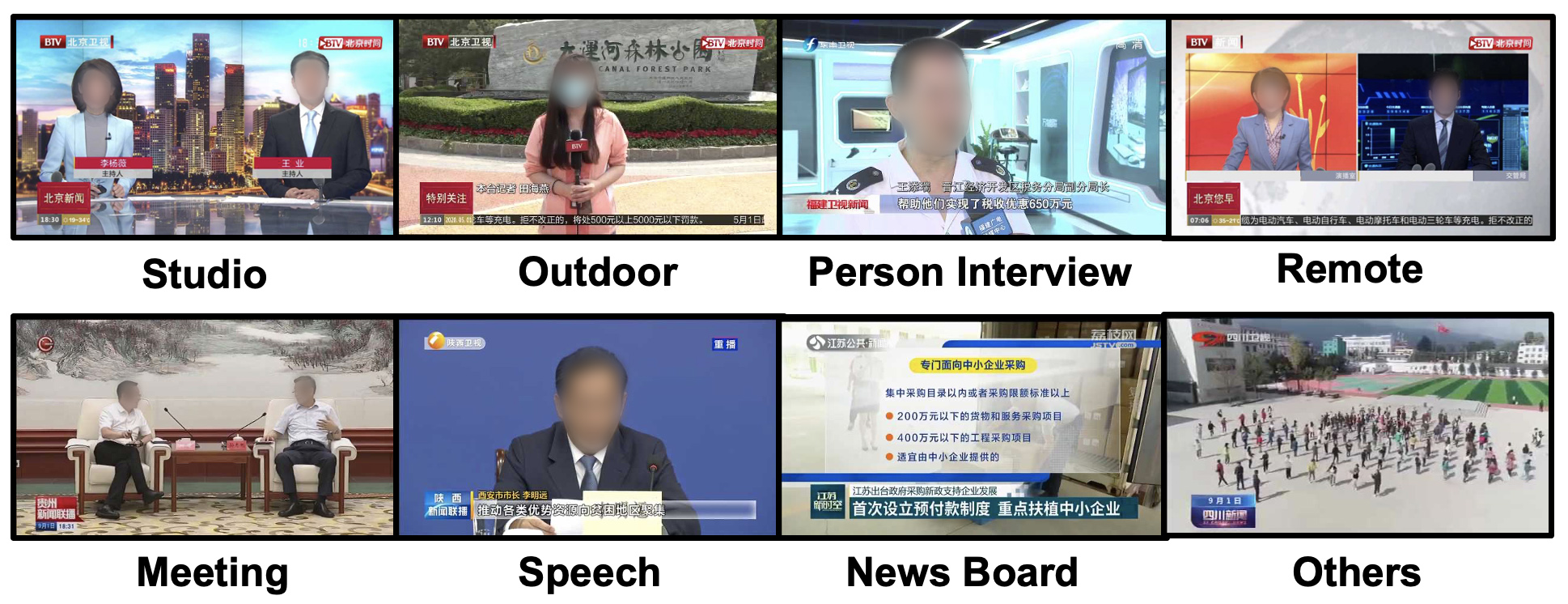}
  \caption{The example images of scene categories in TI-News.}
  \label{FIG:dataset example}
\end{figure}

With carefully analyzing the characteristics of news videos, we established a label system for news SSC task. Seven representative categories are defined. They are \textit{Studio}, \textit{Outdoor}, \textit{Person Interview}, \textit{Remote}, \textit{Meeting}, \textit{Speech} and \textit{News Board}. In addition, class \textit{Others} is used as negative class denoting the scenes which not belong to above-mentioned seven categories. Some graphical illustrations of TI-News dataset is shown in Figure \ref{FIG:dataset example}. In addition, detailed proportions of scene quantities are illustrated in Table \ref{tab:category distribution}.

\begin{table}[]\small
\centering
\caption{The distribution of scene categories.}
\begin{tabular}{l|l|l|l}
\toprule
Studio & Outdoor & Person Interview & Remote \\
\hline
17.8\% & 3.9\%   & 27.7\%  & 1.5\%  \\
\hline
Meeting & Speech & News Board & Others \\
\hline
4.4\%   & 4.9\%  & 1.6\% & 36.7\% \\
\bottomrule
\end{tabular}
\label{tab:category distribution}
\end{table}

TI-News is a large-scale dataset for SS and SSC task and it is the first industrial-grade dataset focusing on news program. Besides the videos selected from various TV stations, the dataset is also equipped with data tools including pre-trained models of unimodal features, video processing tools, visualization tools and etc. Abundant kinds of visual and audio feature extractors are presented, which are pretrained by millions of news videos from short video platform. Therefore, the pre-trained models have strong ability of feature representation for realistic TV shows. 

\noindent \textbf{MovieScenes} MovieScenes is a SS dataset especially for movie. It's frequently used for evaluating the algorithms in scene boundary detection field. The dataset includes 318 movies from a large-scale movie understanding dataset MovieNet \cite{huang2020movienet}. In this paper, to fairly evaluate the scene algorithms, the experimental setup of MovieScenes keeps identical with the work in \cite{rao2020local}.

\subsection{Evaluation Criteria}

According to the definitions of segmentation and classification, the algorithm is evaluated in multiple aspects. Denote the predicted scene segmentation results as $S_{pre}=\left \{b_{1},b_{2},\cdots ,b_{n_{pre}} \right \}$ and classification results as $L_{pre}=\left \{l_{1},l_{2},\cdots ,l_{n_{pre}} \right \}$. The ground truth results are $S_{gt}=\left \{b_{1}^{(gt)},b_{2}^{(gt)},\cdots ,b_{n_{gt}}^{(gt)} \right \}$ and $L_{gt}=\left \{l_{1}^{(gt)},l_{2}^{(gt)},\cdots ,l_{n_{gt}}^{(gt)}\right \}$.

For SS task, the evaluation criteria should measure the difference between $S_{pre}$ and $S_{gt}$. In our experimental settings, the precision and recall of segmentation points (Seg-points) are adopted. For $b_{i}$ in $S_{scene}$, the end shot index $e_{i}$ is denoted as the Seg-point. As a result, the list of Seg-points in $S_{pre}$ is $\{e_{1},e_{2},\cdots,e_{n_{pre}}\}$ and the Seg-points list in $S_{gt}$ is $\{e_{1}^{(gt)},e_{2}^{(gt)},\cdots,e_{n_{gt}}^{(gt)}\}$.

Based on the definition, true positive (TP) is counted by enumerating every predicted and ground truth Seg-points. Concretely, for $\forall b_{i}\in S_{pre}$, if $\exists b_{j}^{(gt)}\in S_{gt}$ satisfy $e_{i}=e_{j}^{(gt)}$, the case will be counted into $TP_{seg}$. The mathematical formulation of $TP_{seg}$ is:

\begin{equation}
TP_{seg} = \sum_{i=1}^{n_{pre}}\sum_{j=1}^{n_{gt}}[[e_{i}=e_{j}^{(gt)}]]
\end{equation}

\noindent where $[[x]]$ is the operator to judge $x$ is true or not, i.e. $[[x]] = 1$ if $x$ is true or 0.

Then, the precision (P), recall (R) and F1 score (F1) for SS can be obtained.

Furthermore, in SSC task, classification is taken into account. Classification aims to assign correct scene label to each segment. A correct predicted scene should have precise scene boundary and accurate label. As a result, the counting of true positive $TP_{seg\&cls}$ will consider category results in addition, which is defined as:

\begin{equation}
TP_{seg\&cls} = \sum_{i=1}^{n_{pre}}\sum_{j=1}^{n_{gt}}[[e_{i}=e_{j}^{(gt)}]]\cdot[[l_{i}=l_{j}^{(gt)}]]
\end{equation}

Similarly, the precision, recall and F1 score for segmentation and classification are calculated. Because of uneven class sizes, micro and macro metrics are both evaluated.

In previous works\citep{chen2021shot,rao2020local}, mean average precision (mAP) and recall are used for evaluate SS. However, the metrics are not consistent and they need enumerating thresholds to calculate while the thresholds are not common in various algorithms. The proposed evaluation criteria doesn't have these limitations and it can directly measure the segmentation and classification outputs. As a result, the proposed criteria is more reasonable and it is used in this paper for fairly methods comparison.


\begin{table*}\small
\centering
\caption{Scene segmentation results on TI-News standard testing set and generalized testing set.}
\begin{tabular}{l|l|l|l|l|l|l} 
\toprule
\multicolumn{1}{l|}{Methods} & \multicolumn{3}{c|}{Standard Testing Set}                                          & \multicolumn{3}{c}{Generalized Testing Set}    \\ 
\hline
& P & R & F1 & P & R & F1  \\ 
\hline
LGSS [Rao et al., 2020]  & 77.68  & 54.94  &  64.36  & 72.96 & 40.62 & 52.19\\ 
\hline
ShotCoL [Chen et al., 2021]  & 78.10  & 67.48  & 72.40  & 75.10 & 53.55 & 62.52\\ 
\hline
LGSS-early [Rao et al., 2020]          & 84.98       & 81.08     & 82.98  & 88.40 & 74.72 & 80.99\\ 
\hline
\hline
\textbf{OS-MSL(SS)}         &   90.14   &  88.70    &     89.41 &     \textbf{90.48} &     87.78 &     89.11                   \\ 
\hline
\hline
Multi-task         &   88.79   &  83.50    &     86.06 &     87.82 &     78.84 &     83.09                           \\ 
\hline
\hline
\textbf{OS-MSL(SSC)}         &   \textbf{90.40}   &  \textbf{89.57}    &     \textbf{89.98} &    89.73 &     \textbf{89.35} &     \textbf{89.54}                           \\ 
\bottomrule
\end{tabular}
\label{tab:SSC experiment 1}
\end{table*}

\begin{table*} \small
  \centering
\caption{Scene segmentation and classification results on TI-News standard testing set and generalized testing set.}
\begin{tabular}{l|l|l|l|l|l|l|l|l|l|l|l|l} 
\toprule
\multicolumn{1}{l|}{Methods} & \multicolumn{6}{c|}{Standard Testing Set}                                          & \multicolumn{6}{c}{Generalized Testing Set}    \\ 
\hline
\multirow{2}{*}{}            & \multicolumn{3}{c|}{Micro} & \multicolumn{3}{c|}{Macro} & \multicolumn{3}{c|}{Micro} & \multicolumn{3}{c}{Macro}  \\ 
\cline{2-13}
& P & R & F1            & P & R & F1    & P & R & F1    & P & R & F1     \\  

\hline
\hline
Two-stage(LGSS-early)                & 77.0 & 73.47 & 75.19 & 69.06 & 65.01 &   66.97  & 77.31 & 65.34 & 70.82 & 53.27 & 43.98 &  48.18         \\
\hline
Two-stage(OS-MSL)                & 83.21  & 81.87 & 82.53 & 76.44  & 75.39  &  75.91  & 83.60 & 81.11 & 82.34 & 59.59 & \textbf{59.56} & 59.57        \\
\hline
Multi-task                 & 81.40 & 76.55 & 78.90   & 80.20 & 64.04  & 71.21    &  79.11  &  71.02  &  74.85  & 57.70 & 43.47 &  48.93  \\
\hline
\hline
\textbf{OS-MSL(SSC)}   & \textbf{86.20}   & \textbf{85.40} & \textbf{85.80}  & \textbf{83.74} & \textbf{78.60} & \textbf{81.09} & \textbf{87.16} & \textbf{86.79} & \textbf{86.97} & \textbf{71.56} & 57.69 & \textbf{63.88}\\
\bottomrule
\end{tabular}
\label{tab:SSC experiment 2}
\end{table*}

\subsection{Compared Methods}



In this paper, the experiments are conducted on both SS task and SSC task. For SS task, \textbf{LGSS} \cite{rao2020local}  and \textbf{ShotCoL} \cite{chen2021shot}  are currently the SOTA algorithms, so we use them as our strong baselines. Other methods such as \textbf{Siamese} \cite{baraldi2015deep}, \textbf{StoryGraph} \cite{tapaswi2014storygraphs}, \textbf{Grouping} \cite{rotman2017optimal} and etc., are not fresh enough to compare. In recent literatures, it is shown that these algorithms perform worse than LGSS to a great extent. Therefore, we conduct comparison experiments with the latest algorithms to reveal the ability of proposed OS-MSL. Note that origin version of LGSS uses late fusion, to utilize the multimodal signals. In this paper, a variant of LGSS is developed with  early fusion. Early fusion means the features of multi modalities are fused before the LSTM in LGSS. The method is denoted as \textbf{LGSS-early}. For SS task, our OS-MSL uses a 5-classes label set $\{$B->I, I->I, I->E, B->E, N$\}$ to address it, which is denoted as \textbf{OS-MSL(SS)}. 

For SSC task, \textbf{Two-stage} and \textbf{Multi-task} are used as baselines. Two-stage consists of two models to address the segmentation and classification problems respectively. Multiple kinds of Two-stage methods can be established by choosing different pairs of segmentation and classification models. In this paper, LGSS-early and our OS-MSL for scene segmentation only are selected as the segmentation model. The corresponding methods are denoted as \textbf{Two-stage (LGSS-early)} and \textbf{Two-stage (OS-MSL)}. On the other hand, \textbf{Multi-task} is built on LGSS-based framework and combined with classification branch to handle the multiple tasks. Detailed models of these methods will be introduced in supplementary material. For SSC task, our OS-MSL uses label set $C \times \{$B->I, I->I, I->E, B->E, N$\}$ to address 
it, which is denoted as \textbf{OS-MSL(SSC)}.



\subsection{Results on TI-News Dataset}

In this subsection, OS-MSL is evaluated on TI-News dataset with standard testing set and generalized testing set. The standard testing set refers to 34 videos from 8 programs which are included in training set. On the contrary, in generalized testing set, 10 more testing videos are selected from distinct news programs and their formats of program editing are very different with training samples. Therefore, the generalized testing set can evaluate the generalization of scene methods.


The results of SS experiments are shown in Table \ref{tab:SSC experiment 1}. Our OS-MSL(SS) outperforms the strong baselines, i.e. LGSS, LGSS-early and ShotCoL. It reflects that the scheme of Sequential Link is superior for solving the segmentation problem. To explore the reasons, we summarize that OS-MSL(SS) reforms SS task to a link form, and explicitly reveal richer sequential relation between shots using a 5-classes link label set. In generalized testing set, OS-MSL(SS) also shows its splendid performance. It indicates OS-MSL(SS) has great generalization ability on SS task. Furthermore, the SSC methods, Multi-task and OS-MSL(SSC), are examined their ability of SS. As the Table \ref{tab:SSC experiment 1} shows, Multi-task outperforms LGSS-early and OS-MSL(SSC) achieves the best results, which illustrate that utilizing scene category information can promote SS task.


SSC results are illustrated in Table \ref{tab:SSC experiment 2}. The micro precision and micro recall of OS-MSL(SSC) reach $86.20\%$ and $85.40\%$ for the evaluation of segmentation and classification. Both precision and recall exceed $85\%$, which lead to an excellent performance of SSC task. In supplementary material, some visualized examples are presented. It can be found that the visualization results are nearly in accord with the human's feeling. Compared with Two-stage and Multi-task, proposed OS-MSL(SSC) shows great superiority. It is verified that the Sequential Link framework can better exploit the relation between segmentation and classification tasks. 




In generalized testing set, for OS-MSL(SSC), micro F1 score of generalized videos exceed $86\%$. It illustrates the good generalization of our algorithm. Relatively, the performance of LGSS, ShotCoL, Two-stage and Multi-task decreases to a large extent. These phenomenons further support the superiority of our method. 

However, the results are quite bad under the macro evaluation. The macro F1 scores of these methods are in the range of $48.93\%\sim63.88\%$. The reason is that the algorithms perform poorly to find \textit{Outdoor}, \textit{Remote}, \textit{Speech} categories, which greatly affect the macro results. The generalization ability of recognizing hard categories is a point for improvement.


\begin{table} \small
\centering
\caption{Scene segmentation results on MovieScenes.}
\begin{tabular}{l|l|l|l} 
\toprule
Methods              & \multicolumn{3}{c}{Segmentation}  \\ 
\hline
                     & P & R & F1            \\ 
\hline
LGSS-early [Rao et al., 2020]        &  45.56  & 44.97  & 45.26        \\ 
\hline
LGSS [Rao et al., 2020]        & 45.64     & 46.36  & 46.00         \\ 
\hline
ShotCoL [Chen et al., 2021]         & \textbf{51.02}     & 47.97  & 49.45         \\ 
\hline
\hline
\textbf{OS-MSL(SS)}     & 49.89     & \textbf{50.56} & \textbf{50.22}   \\
\bottomrule
\end{tabular}
\label{tab:SSC experiment 3}
\end{table}

\begin{table}\small
\centering
\caption{Ablation study on TI-News standard testing set.}
\begin{tabular}{l|l|l|l|l|l|l} 
\toprule
\multicolumn{1}{l|}{Methods} & \multicolumn{6}{c}{Segmentation and Classification}    \\ 
\hline
\multirow{2}{*}{}       & \multicolumn{3}{c|}{Micro} & \multicolumn{3}{c}{Macro}  \\ 
\cline{2-7}
                             &   P & R & F1    & P & R & F1     \\ 
\hline
\textbf{OS-MSL(SSC)}   & 86.20 & \textbf{85.40} & \textbf{85.80}  & \textbf{83.74} & \textbf{78.60} & \textbf{81.09}  \\
\hline
\hline
w/o Audio    & 83.26 & 78.50 & 80.81 & 76.50 & 70.41 & 73.33 \\ 
\hline
w/o Visual    & 81.57 & 71.36 & 76.12 & 69.44 & 40.59 & 51.24 \\ 
\hline
\hline
w/o BB Training  & 85.86   & 81.90  & 83.83 & 82.09 & 73.07 & 77.32       \\ 
\hline
w/o BN         & 85.27    & 79.95  & 82.52 & 80.24 & 69.09 & 74.25            \\ 
\hline
w/o DiffCorrNet  & \textbf{86.24}  & 79.63   & 82.80   & 83.36 & 71.61 & 77.04   \\ 
\hline
w/o CRF     & 84.60  & 82.11  & 83.34 & 82.65 & 74.17 & 78.18 \\ 
\bottomrule
\end{tabular}
\label{tab:Ablation experiment standard}
\end{table}

\begin{table} \small
\centering
\caption{Ablation study on TI-News generalized testing set.}
\begin{tabular}{l|l|l|l|l|l|l} 
\toprule
\multicolumn{1}{l|}{Methods} & \multicolumn{6}{c}{Segmentation and Classification}    \\ 
\hline
\multirow{2}{*}{}       & \multicolumn{3}{c|}{Micro} & \multicolumn{3}{c}{Macro}  \\ 
\cline{2-7}
                             &  P & R & F1    & P & R & F1     \\
\hline
\textbf{OS-MSL(SSC)}   & \textbf{87.16}  & \textbf{86.79} & \textbf{86.97}  & \textbf{71.56} & 57.69 & \textbf{63.88}\\
\hline
\hline
w/o Audio  & 78.39 & 69.03 & 73.41 & 54.55 & 47.37 & 50.70 \\ 
\hline
w/o Visual  & 82.76 & 71.59 & 76.77 & 57.55 & 32.95 & 41.91 \\ 
\hline
\hline
w/o BB Training  & 85.56  & 78.27  & 81.75  & 60.54 & 52.82 & 56.42 \\ 
\hline
w/o BN    & 86.40    & 76.70  & 81.26  & 63.98 & 53.55 & 58.30  \\ 
\hline
w/o DiffCorrNet  & 86.20 & 78.98 & 82.43   & 67.39 & \textbf{60.41} & 63.71   \\ 
\hline
w/o CRF  & 83.46 & 80.26 & 81.83 & 66.59 & 56.76 & 61.29 \\ 


\bottomrule
\end{tabular}
\label{tab:Ablation experiment generalized}
\vspace{-4ex}
\end{table}

\subsection{Results on MovieScenes Dataset}

The experiments on public available data set of MovieScenes are conducted to evaluate our performance. MovieScenes is established just for SS task so that only segmentation performance is evaluated. 


In Table \ref{tab:SSC experiment 3}, it illustrates our OS-MSL(SS) outperforms the latest algorithms, i.e. LGSS, LGSS-early and ShotCoL. As a result, the scheme of our Sequential Link is also effective on movie domain. It can further support that the proposed Sequential Link framework is superior for solving the segmentation problem alone.





\subsection{Ablation Study}

To further explore the effectiveness of several modules designed in OS-MSL, the ablation study experiments of 1) \textbf{w/o Backbone (BB) Training}, 2) \textbf{w/o BN}, 3) \textbf{w/o DiffCorrNet} and 4) \textbf{w/o CRF} are conducted. Table \ref{tab:Ablation experiment standard} shows the ablation results of SSC task on TI-News standard testing set. OS-MSL(SSC) w/o BB Training, w/o BN and w/o DiffCorrNet all perform worse F1 score compared with OS-MSL(SSC). The phenomenon illustrates all these modules have their positive effects for SSC to different extent. The ablation study of OS-MSL(SS) on TI-News has the similar conclusion. Its detailed experiments are described in supplementary material.

It is feasible to use other sequence model to tag shots. However, CRF is an effective module because it can utilize the transition relations between sequence labels. For example, it's impossible that a ${Meeting}_{B->I}$ follows a ${Meeting}_{I->I}$. As Table \ref{tab:Ablation experiment standard} shows, OS-MSL(SSC) w/o CRF performs $1.6\%/3.3\%$ on precision/recall lower than proposed OS-MSL(SSC). In addition, we also substituted transformer with LSTM to explore the sequence model selection. The model achieves $83.87\%/78.63\%$ micro precision and recall, which is much worse than the results of transformer. The experiments illustrate the model of transformer with CRF is the better choice.

In Table \ref{tab:Ablation experiment generalized} and Table \ref{tab:Ablation experiment MovieScenes}, the ablation experiments on TI-News generalized testing set and MovieScenes testing set are conducted. The results also reflect that all the modules are effective to improve the performance of SS and SSC task.

\begin{table} \small
\centering
\caption{Ablation study on MovieScenes.}
\begin{tabular}{l|l|l|l} 
\toprule
Methods              & \multicolumn{3}{c}{Segmentation}  \\ 
\hline
                     & P & R & F1            \\ 
\hline
\textbf{OS-MSL(SS)}     & 49.89     & \textbf{50.56} & \textbf{50.22}   \\
\hline
\hline
w/o Audio    & \textbf{51.06} & 46.59 & 48.72 \\ 
\hline
w/o Visual    & 28.57 & 13.51 & 18.34 \\ 
\hline
\hline
w/o BB Training  & 49.73 & 49.60 & 49.67 \\ 
\hline
w/o BN         & 49.25 & 50.38 & 49.81  \\ 
\hline
w/o DiffCorrNet  & 47.94 & 47.83 & 47.89 \\ 
\hline
w/o CRF     & 48.15  & 50.43  & 49.27  \\ 

\bottomrule
\end{tabular}
\label{tab:Ablation experiment MovieScenes}
\end{table}

\section{Discussion}
\subsection{Multimodal Fusion}

In this subsection, we try to explore the effects of different modalities on the performance.Firstly, in news video scenario, as Table \ref{tab:Ablation experiment standard} and \ref{tab:Ablation experiment generalized} show, our proposed multimodal method outperforms either unimodality method. In fact, audio and visual modalities have their specific roles, which should be complementary. Audio contains strong clues for segmentation because the switch of voices tends to be the Seg-points. However, classification only with audio modality is almost infeasible. Some scenes categories (such as \textit{Studio}, \textit{Meeting} and \textit{News Board}) can not be distinguished because most of them have the same voice. As a result, the experimental results of \textbf{OS-MSL w/o Visual} are quite low. On the other hand, visual unimodality is also insufficient. The visual expression has large inner-class variance. For example, the frames of \textit{Person Interview} include not only person talking but also the things describing what the person is talking about. The Seg-points of this case are hard to determine.

Secondly, in movie scenario, as Table \ref{tab:Ablation experiment MovieScenes} shows, \textbf{OS-MSL w/o Visual} performs poorly in MovieScenes dataset. On the contrary, \textbf{OS-MSL w/o Audio} achieves decent results, which are slightly worse than \textbf{OS-MSL}. The phenomenons illustrate that the audio signals are insufficient and visual modality are dominated to segment movie scenes. Nevertheless, audio modality also contains its exclusive clues to assist the task, which makes the multimodal method outperform the unimodal ones.

It may be argued that text is another useful modality. With the techniques of optical character recognition (OCR) and automatic speech recognition (ASR), text content in frames and audio can be recognized and offers the detail semantic information about the video. However, for the SSC task, the text modality is almost useless. The text content is little related to the definition of SSC task and sometimes it is noise which may cause negative effect.

Some cases may support this opinion. In news video scenario, presenter may report several news in the studio, but these news are unrelated to \textit{Studio} category and we also can't judge the scene boundary by the content of these news. In movie scenario, the words of background music or voice are meaningless. As a result, we abandon the text information and just use visual and audio modalities to address SSC task.

\subsection{Learning Curves}

\begin{figure}[t]
  \centering
  \includegraphics[width=0.7\linewidth]{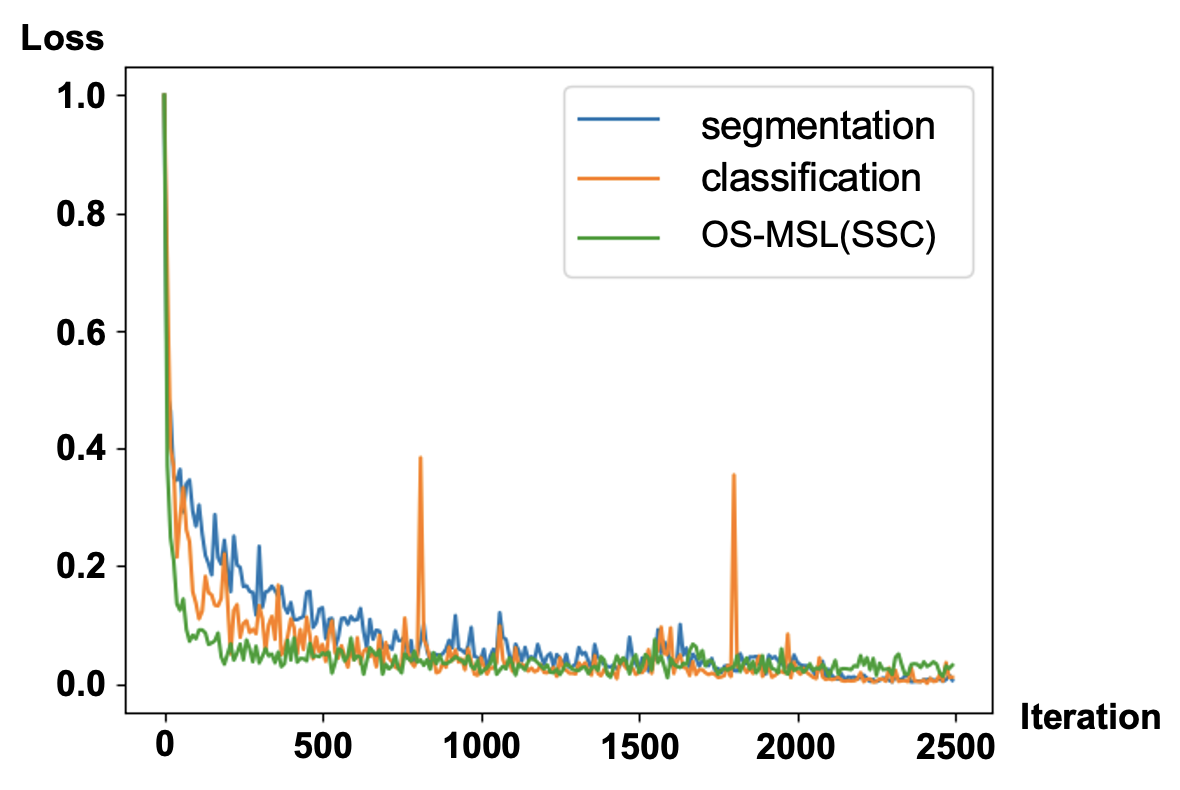}
  \caption{Comparison of learning curve between multi-task and OS-MSL(SSC).}
  \label{FIG:learning curve}
\end{figure}

In this subsection, we focus on the model collapse problem. During the training phase, the \textbf{segmentation} loss and \textbf{classification} loss of multi-task method are recorded individually. For comparison, the learning curve of OS-MSL(SSC) is also plotted, which is denoted as \textbf{OS-MSL(SSC)}. Each curve is drawn according to the losses after normalization due to the different magnitudes of various tasks.

In Figure \ref{FIG:learning curve}, it is illustrated that the learning curve of classification oscillates greatly. The reason is that the model is dominated by segmentation task. The gradients during training tend to optimize the segmentation task, which is inconsistent with classification task. As a result, the conflict between the two tasks may affect the model training, which finally harms the performance. Comparatively, the learning curve of OS-MSL(SSC) converges more rapidly and is more stable. The phenomenon reflects our one-stage method can alleviate the asynchronization issue existing in multi-task learning and obtain better performance for SSC task.

\subsection{Sequential Link Framework}

The tasks of SS and SSC can be characterized in a unified manner: a link form with link tagging. Furthermore, our framework can be easily generalized to other 
similar video sequential modeling tasks, e.g., action, activity or event
recognition, which can be categorized to a unified definition: find a boundary between  consecutive segments/clips with auxiliary information such as label, key frame, and caption. For example,  video action recognition aims to recognize human actions in a video, thus the task is to find  
consecutive video frames with the same action labels, which can be obviously completed by our framework. 

\section{Conclusion}

The paper raises a novel video structuring problem, i.e. SSC. To address the model collapse problem existing in two-stage and multi-task methods, Sequential Link framework is introduced to reform the two tasks into a unified one using a novel link form. Based on this, OS-MSL is proposed to leverage common information and avoid disharmony of the two tasks. To strength the shot representation, a new module DiffCorrNet is developed to extract differences and correlations among successive shots. For news program application, a specific dataset, TI-News, is established. The experiments on TI-News and MovieScenes illustrate its effectiveness.

\bibliographystyle{ACM-Reference-Format}
\bibliography{sample-base}

\end{document}